\documentclass{article}

\usepackage{PRIMEarxiv}

\usepackage[utf8]{inputenc} 
\usepackage[T1]{fontenc}    
\usepackage{booktabs}       
\usepackage{amsfonts}       
\usepackage{nicefrac}       
\usepackage{microtype}      
\usepackage{lipsum}
\usepackage{fancyhdr}       
\usepackage{graphicx}       
\graphicspath{{media/}}     

\usepackage[english]{babel}
\usepackage[utf8]{inputenc}
\usepackage{amsmath}
\usepackage{tikz,pgfplots}
\usepackage{tikz-network}
\usetikzlibrary{shapes,decorations,arrows,calc,arrows.meta,fit,positioning}
	
\usepackage{color, colortbl}
	
\definecolor{Gray}{gray}{0.9}

\title{Evaluating Counterfactual Explanations Using Pearl's Counterfactual Method
}

\author{
  Bevan I. Smith \\
  School of Mechanical, Industrial and Aeronautical Engineering,\\
  University of the Witwatersrand, \\
  Johannesburg, South Africa\\
  \texttt{bevan.smith@wits.ac.za} \\
}

\begin{document}
\maketitle

\begin{abstract}
Counterfactual explanations (CEs) are methods for generating an alternative scenario that produces a different desirable outcome.  For example, if a student is predicted to fail a course, then counterfactual explanations can provide the student with alternate ways so that they would be predicted to pass.   The applications are many.  However, CEs are currently generated from machine learning models that do not necessarily take into account the true causal structure in the data.  By doing this, bias can be introduced into the CE quantities. I propose in this study to test the CEs using Judea Pearl's method of computing counterfactuals which has thus far, surprisingly, not been seen in the counterfactual explanation (CE) literature.   I furthermore evaluate these CEs on three different causal structures to show how the true underlying causal structure affects the CEs that are generated.  This study presented a method of evaluating CEs using Pearl's method and it showed, (although using a limited sample size), that thirty percent of the CEs conflicted with those computed by Pearl's method.  This shows that we cannot simply trust CEs and it is vital for us to know the true causal structure before we blindly compute counterfactuals using the original machine learning model.  
\end{abstract}


\section{Background}
This study presents a method for validating counterfactual explanations using Pearl's counterfactual method \cite{Pearl2016, Pearl2017}. Counterfactual explanations (CEs) form part of the rapidly expanding field of explainable machine learning which aims to explain why machine learning models make their predictions \cite{Molnar2019, Wachter2017}.  If the predictions made by the model were undesirable, knowing why the predictions were made would allow us to generate counterfactual explanations that give us alternative desirable outcomes. For example, if a student was predicted to fail, we could generate CEs that would advise the student how to change certain features to increase her probability of passing. A counterfactual explanation (CE) ``\textit{describes the smallest change to the feature values that changes the prediction to a predefined output}" \cite{Molnar2019}.  Clearly we can see that CEs promise much and the applications are vast, such as improving pass rates, advising patients what to change to improve health, advising clients what to do so they can obtain a bank loan, and so forth.  
\vspace{0.2cm}
\noindent The main problem however is that CEs, and the algorithms that generate them, are yet to be properly validated \cite{Verma2021}. Why would they need to be validated or tested?  This is because CEs generated using the various optimization algorithms such as DiCE and WACH \cite{Mothilal2020,Wachter2017}, are generated using the model trained on the original data \cite{Molnar2019}.  This is problematic because machine learning (ML) models do not care about the causal relationships and causal structure between the features,  only correlation \cite{Pearl2016}.  The ML model is designed to make good predictions without concern for causality.  \textit{Therefore if the algorithms that generate CEs are based on these ML models, would they be the same as those generated via a ground truth structural causal model?}  We already know that when we fit simple linear regression models without taking into account the causal structure, there are biases in the coefficients \cite{Luque-Fernandes2018}.  Examples include leaving out a confounding feature in a regression model and including a collider in the model.  If not taking into account causal relationships in a regression model causes problems, how much more if we generate counterfactual explanations using the existing trained ML model.  The problem that this study is aiming to address is the following:  Counterfactual explanations are generated using machine learning models that do not necessarily take into account the true underlying causal structure in the data.  This can lead to erroneous estimations of the counterfactuals.

\vspace{0.2cm}
\noindent To address this problem, I propose using the counterfactual methods found in the work of Judea Pearl\cite{Pearl2016,Pearl2017}.  Details about this method is found later in Section \ref{PCM}.  However, the essential idea is that to compute counterfactuals, we first require a structural causal model (SCM) and associated graph that describe the true data generating process.  This allows us to compute counterfactuals using the true causal relationships between the data, whether it be between input features or between input and output features.   Whereas computing CEs from machine learning models would not necessarily incorporate causal relationships, Pearl's method does.   I propose using this method to evaluate and test the current CE methods.

The main contributions in this paper are:

\begin{itemize}
    \item I present a method for evaluating CEs using the counterfactual method developed by Pearl.  This has surprisingly not been applied in the CE literature thus far.  This would allow us to evaluate our CEs to know if we can trust them.
    \item Using this method, I show how the CEs generated using current methods conflict with those computed via Pearl's method and how different causal structures affect the results.  This should alert us to the fact that it is vital to understand the true underlying causal structure before generating CEs.
\end{itemize}

\section{Counterfactual Explanations }
\label{CounterfactualExplanations}
Recently, counterfactual explanations (CEs) have been gaining much traction and interest in the machine learning literature \cite{Wachter2017,Molnar2019,Mahajan2020, Verma2021}.  The aim of CEs is to manipulate the inputs of a model to change the output to a desired one.  CEs are generated by solving an optimization problem where the input feature space is perturbed to as close a feature space as possible, but one that leads to a different, desirable output \cite{Mahajan2020, Molnar2019}.  This is therefore an optimization problem that aims to perturb the input space as little as possible, in order to change the output.  

\subsection{Method by Wachter et al.}

An example of a CE method is that by Wachter et al. \cite{Wachter2017}.  The loss function is given below:

\begin{equation}
   L(x,x',y',\lambda) = \lambda \cdot(f(x') - y')^2 + d(x,x'),
\end{equation}

\noindent where $x$ and $x'$ are the original and counterfactual input space respectively, $y'$ is the counterfactual output, and $\lambda$ a tuning parameter. Importantly, $f$ refers to the original machine learning model trained on the data, which does not necessarily take into account causal relationships in the data.  $L$ aims to balance between minimizing the distance between original and counterfactual $x$ and $x'$ and predicted $f(x')$ and original $y'$ outputs.

\subsection{Method by Mothilal et al.}
Another CE method was developed by Mothilal et al. \cite{Mothilal2020} and is the method used in this study.  It is known as DiCE: \textbf{Di}verse \textbf{C}ounterfactual \textbf{E}xplanations.  Their method of optimization is based on the following loss function:

\begin{equation}
  C(x) = \underset{{c_1,...,c_k}}{\arg\min}\frac{1}{k} \sum_{i=1}^k yloss(f(c_i),y) + \frac{\lambda_1}{k}\sum_{i=1}^{k}dist(c_i,x) - \lambda_2  dpp\_diversity(c_1,...,c_k),  \label{dice_equation}
\end{equation}

\noindent where $c_i$ is a counterfactual explanation (CE), \textit{k} is the total number of CEs, \textit{f}(.) refers to the ML model trained on the data, \textit{yloss}(.) is a metric minimizing the distance between counterfactual predictions and the desired outcome, \textit{dist}(.) is the distance between counterfactual input $c_i$ and actual input \textit{x}, \textit{dpp\_diversity}(.) functions to maximize diversity, and finally the lambda values balance the three parts of the loss function. 
\noindent An important aspect of this study is to show that CE methods shown above use the ML model trained on the data.  

\subsection{Characteristics of CEs}
CEs require important characteristics.  I highlight the following:

\begin{itemize}
    \item Feasibility/plausibility:  This refers to the CEs being realistic; not being smaller or larger than those observed in the original data \cite{guidotti22}.  It means the CEs must come from a possible world \cite{Wachter2017}.
    \item Actionable:  Features must be able to be modified, to be changed.  Features that are non-actionable include age, gender, race etc. \cite{guidotti22}.
    \item Causality:  I discuss this in more detail in the Discussion section (Section \ref{discussion}).  This characteristic requires that causal relationships be maintained in the CEs.  However, I push back on this requirement.
    
\end{itemize}

\section{Structural Causal Models}
\label{structuralcausalmodels}
\begin{figure}[h!]
    \centering
    \includegraphics[width = 15cm]{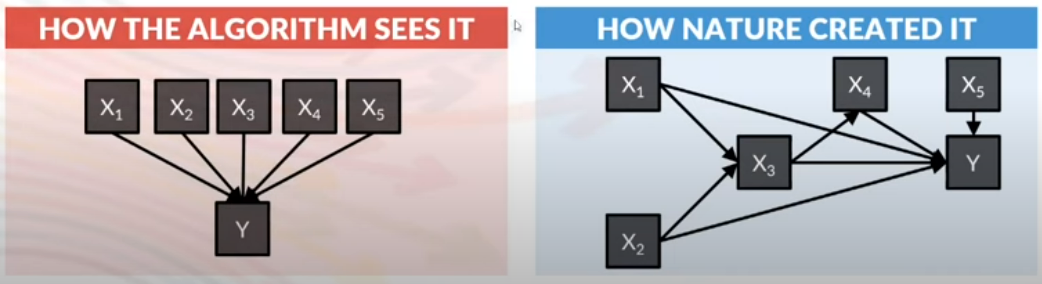}
    \caption{ML model (left) vs SCM (right) \cite{peter_tennant_vid}.}
    \label{algvsscm}
\end{figure}

 Structural causal models (SCMs) refer to the true data generation process, and allow for counterfactual analysis \cite{Verma2021, Pearl2016}.  SCMs are associated with graphical causal models called directed acyclic graphs (DAGs), seen on the right side of Figure \ref{algvsscm}. Consider a set of features $X_1, ..., X_n$, known as endogenous features, where each feature $X_i$ is generated via a deterministic function $f_i$ which is a function of its causes (or parents, \textit{PA}). $PA_i$ refers to the parents (known causes) of $X_i$,. and $U_i$ refers to some stochastic unexplained cause of $X_i$ \cite{Scholkopf2019}, also known as exogenous features (outside the model). Each feature corresponds to a vertex in the DAG. 
 
 \begin{equation}
     X_i:= f_i(PA_i, U_i)~~~~~~~(i = 1,...,n)
 \end{equation}
 
Figure \ref{algvsscm} presents some intuition behind the difference between generating counterfactuals based on a machine learning model (left image) and a true structural causal model DAG (right image) \cite{peter_tennant_vid}.  SCMs would generate counterfactuals based on the true causal structure but CEs are generated based on the original machine learning model $f$ which does not necessarily take into account causal structure. The question is: \textit{would the CE algorithms generate the same counterfactuals as those via SCMs}?

\section{Causal Structures}
\label{causalstructures}
Before presenting Pearl's Counterfactual Method (PCM) in Section \ref{PCM}, I present three important causal structures in causal inference: chains (or mediators), forks (common cause) and colliders.  This study will use these three structures when evaluating the CEs using PCM.  According to Pearl, these three types are the building blocks of causal structures and ``enable us to test a causal model, discover new models, evaluate effects of interventions, and much more" \cite{Pearl2017}.  They are presented using a directed acyclic graph and an SCM.

\subsection{Chain (mediator)}
\label{causalstructurechain}
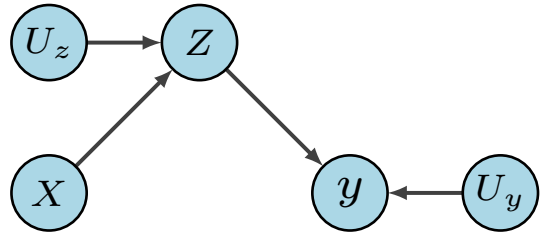
\begin{figure}[h!]
\centering
\begin{tikzpicture}
\Vertex[x=1,size = 1,label = $X$, fontscale = 2]{X}
\Vertex[size =1, x=5,label = $y$, fontscale =2.5]{Y}
\Vertex[size=1,x=3,y=2,fontscale = 2,label = $Z$]{Z}
\Vertex[size=1,x=7,y=0,label = $U_y$,fontscale = 2]{Uy}
\Vertex[size=1,x=1,y=2,label = $U_z$,fontscale = 2]{Uz}
\Edge[Direct](Uz)(Z)
\Edge[Direct](Uy)(Y)
\Edge[Direct](X)(Z)
\Edge[Direct](Z)(Y)
\end{tikzpicture}
\caption{DAG showing chain path between \textit{X} and \textit{Z} and \textit{y}. Causality flow from \textit{X} to \textit{Z} to \textit{y}. }
\label{chain1}
\end{figure}

The chain or mediator DAG is shown in Figure \ref{chain1}. An example of a chain would be a fire (\textit{X}), smoke (\textit{Z}) and alarm (\textit{y}) system.  The fire does not directly cause the alarm to go off, but is required to first produce smoke that then sets off the alarm.  There is no direct causal path between fire and alarm, in this case.  Of course, there can be a causal path between \textit{X} and \textit{y} in other cases.  We are restricting ourselves here to the basic chain. What this model tells us is that if we control for \textit{Z}, then we block the causal flow between \textit{X} and \textit{y} and do not allow for measuring the true effect of \textit{X} on \textit{y}.  That is, \textit{X} and \textit{y} are conditionally independent given \textit{Z}. What this implies practically is that if we include \textit{X}, \textit{Z} and \textit{y} in our model, we are effectively controlling for \textit{Z} and blocking the path between \textit{X} and \textit{y}.  The SCM for Figure \ref{chain1} is seen in Equations \ref{eq15} and \ref{eq16}. Here we introduce the exogenous variables, \textit{U}, that are vital to Pearl's method seen later. \textit{Z} is a function of \textit{X} and $U_z$, where \textit{X} is an endogenous feature (e.g. fire) and $U_z$ refers to all external features causing \textit{Z} that we can not account for one by one. This goes for all \textit{U} features seen later.

\vspace{1cm}

\begin{align}
&Z = f(X, U_z) \label{eq15}\\
&y = f(Z,U_y)   
\label{eq16}
\end{align}

\subsection{Fork (common cause)}
\label{common parent- section}
\begin{figure}[h!]
\centering
\begin{tikzpicture}
\Vertex[x=1,size = 1,label = $X$, fontscale = 2]{X}
\Vertex[x=-1,size = 1,label = $U_x$, fontscale = 2]{Ux}
\Vertex[size =1, x=5,label = $y$, fontscale =2.5]{Y}
\Vertex[size=1,x=3,y=2,fontscale = 2,label = $Z$]{Z}
\Vertex[size=1,x=7,y=0,label = $U_y$,fontscale = 2]{Uy}
\Vertex[size=1,x=1,y=2,label = $U_z$,fontscale = 2]{Uz}
\Edge[Direct](Uz)(Z)
\Edge[Direct](Ux)(X)
\Edge[Direct](Uy)(Y)
\Edge[Direct](Z)(X)
\Edge[Direct](Z)(Y)
\end{tikzpicture}
\caption{DAG showing common parent confounding where \textit{Z} is the parent of both \textit{X} and \textit{y}, but with no causal flow from \textit{X} to \textit{y} in this case.  Also known as spurious correlation.}
\label{conf2}
\end{figure}
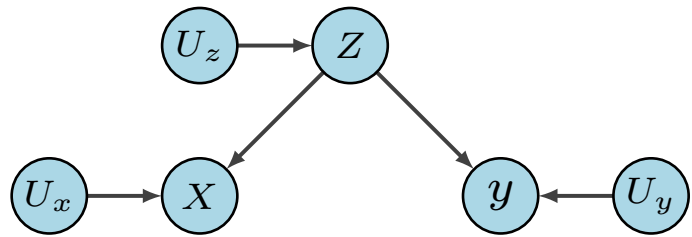

Figure \ref{conf2} shows a DAG representing a fork or common cause causal structure.  Practically, it represents spurious correlation where \textit{X} and \textit{y} are conditionally independent given \textit{Z}. That is, in this case, to obtain the true causal effect of \textit{X} on \textit{y} requires conditioning (controlling) on \textit{Z}. The SCM for Figure \ref{conf2} is shown in Equations \ref{eq12} to \ref{eq13}.  An example of this particular causal structure could be where \textit{X} refers to shark attacks and \textit{y} refers to ice-cream sales.  We may obtain data that shows that these two events are correlated:  as shark attacks increase, ice-cream sales also increase.  However, what explains this spurious correlation is \textit{Z}, which refers to summertime.  Both shark attacks and ice-cream sales are conditionally dependent on the seasons. 



\vspace{2cm}

\begin{align}
&X = f(Ux) \label{eq12}\\
&Z = f(Uz) \\
&y = f(Z,Uy)   \label{eq13}
\end{align}    

\subsection{Collider}
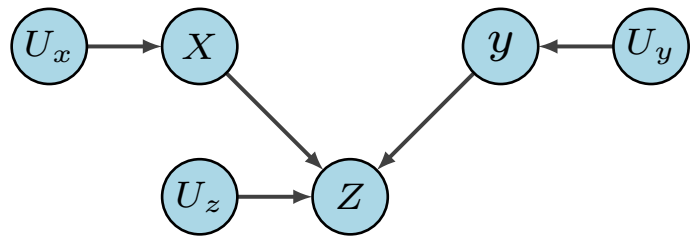
\begin{figure}[h!]
\centering
\begin{tikzpicture}
\Vertex[x=1,size = 1,label = $X$, fontscale = 2]{X}
\Vertex[x=-1,size = 1,label = $U_x$, fontscale = 2]{Ux}
\Vertex[size =1, x=5,label = $y$, fontscale =2.5]{Y}
\Vertex[size=1,x=3,y=-2,fontscale = 2,label = $Z$]{Z}
\Vertex[size=1,x=7,y=0,label = $U_y$,fontscale = 2]{Uy}
\Vertex[size=1,x=1,y=-2,label = $U_z$,fontscale = 2]{Uz}
\Edge[Direct](Uz)(Z)
\Edge[Direct](Uy)(Y)
\Edge[Direct](X)(Z)
\Edge[Direct](Y)(Z)
\Edge[Direct](Ux)(X)
\end{tikzpicture}
\caption{DAG showing collider path between \textit{X}, \textit{Z} and \textit{y}, but with no causal path between \textit{X} and \textit{y}.}
\label{coll1}
\end{figure}

The final causal structure used in this study is the collider, presented by the DAG in Figure \ref{coll1} with no causal path between \textit{X} and \textit{y}. An example of a collider is where \textit{X} is a sprinkler, \textit{y} is rain, and \textit{Z} is wet grass. Each feature can be on or off, so to speak.  \textit{Z} is therefore a function of the sprinkler and rain.  For example, if \textit{Z} = dry (off), then both sprinkler (\textit{X}) and rain (\textit{y}) must be off.  However, if \textit{Z} = wet, either \textit{X} or \textit{y} is on, or both are on.  The important point to note here is that \textit{X} and \textit{y} are marginally independent; they are independent, conditioned on nothing.  Rain being on has nothing to do with the sprinkler being on, for example.  However, when we condition on \textit{Z}, we make sprinkler and rain conditionally dependent.  This means that by conditioning on grass, we create the illusion that there is a causal relationship between otherwise independent features. For example, if we know that grass is wet (\textit{Z} = on) and that it is raining (\textit{y} = on), then we immediately know the value of the sprinkler (\textit{X} = off).  Two variables (\textit{X} and \textit{y}) that are independent, become conditionally dependent, given \textit{z}.  This is the opposite of chains and forks where conditioning on \textit{Z} results in estimating the actual causal relationship between \textit{X} and \textit{y}.  The SCMs for the collider is seen in Equations \ref{eq19} to \ref{eq20}.

\vspace{2cm}


\begin{align}
&Z = f(X, y, U_z) \label{eq19}\\
&X = f(U_x)
&y = f(U_y   )   \label{eq20}
\end{align}


\subsection{Difference between SCM and ML}

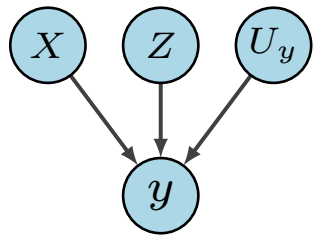
\begin{figure}[h!]
\centering
\begin{tikzpicture}
\Vertex[x=1,size = 1,y=2,label = $X$, fontscale = 2]{X}
\Vertex[size =1, x=2.5,label = $y$, fontscale =2.5]{Y}
\Vertex[size=1,x=2.5,y=2,fontscale = 2,label = $Z$]{Z}
\Vertex[size=1,x=4,y=2,label = $U_y$,fontscale = 2]{Uy}
\Edge[Direct](Uy)(Y)
\Edge[Direct](X)(Y)
\Edge[Direct](Z)(Y)
\end{tikzpicture}
\caption{DAG showing how a supervised machine learning model would train on the data.}
\label{ml1}
\end{figure}

\noindent What is vital to note from these three basic causal structures is that whereas the SCMs would take into account the causal structure (i.e. the data generating process), the ML model samples from the joint distribution of \textit{X}, \textit{y} and \textit{Z} without any thought for the causal structure. Clearly these three types are distinctly different.  For example, the DAG and linear model for an ML model might look like that shown in Figure \ref{ml1} and Equation \ref{eq14}. The question is, if we generate CEs from the ML model, would they be the same as those generated by the true SCM using Pearl's method shown next?

\vspace{0.2cm}

\newpage
\begin{align}
y = f(X,Z,Uy)   \label{eq14}
\end{align}

\section{Pearl's Counterfactual Method (PCM)}
\label{PCM}
We now come to Pearl's Counterfactual Method (PCM).  This sections presents how we go about generating counterfactuals using SCMs.  The method is based on Judea Pearl's work that can be found in Chapter 8 of The Book of Why \cite{Pearl2017} and Chapter 4 of Causal Inference in Statistics, A Primer \cite{Pearl2016}.  The method involves three steps: abduction, action and prediction and works as follows.    
\subsection{Abduction: to compute exogenous variables, \textit{U}}
\label{abduction}
Consider data with a known true causal structure.  This structure has a DAG and SCMs. For example, consider the DAG and SCM for Figure \ref{conf2} where we have spurious correlation from a common parent.  The SCMs are given as \textit{Z = f(Uz)} and \textit{y = f(Z,Uy)}.  The initial aim in PCM is to compute the values of the exogenous variables \textit{Uz} and \textit{Uy}.  These refer to all unmeasured variables \textit{for an individual unit}, that affect \textit{Z} and \textit{y}, respectively.  These \textit{U} variables, also referred to as noise variables, describe the world of an individual person or observation.  It is these that we need to compute first before perturbing the input features to compute counterfactuals.  We need to compute these variables first because they describe the "situation" of an individual unit that must remain the same in both the factual and counterfactual worlds. Recall that the idea behind the counterfactual is to keep all other things constant and to change only one feature. These other things refer to the exogenous variables. 

To compute \textit{Uz} and \textit{Uy} in our example, we select a unit (eg. a patient, a student etc.) out of the data that we are interested in, and input \textit{observed factual features} into our SCMs. For example, say we measured \textit{X}, \textit{Z} and \textit{y} from observation.  We input the individual unit's \textit{Z} into the first SCM, \textit{Z = f(Uz)}, and compute \textit{Uz}.  To compute \textit{Uy}, input measured \textit{y} and \textit{Z} from the individual unit, into the second SCM, \textit{y = f(Z,Uy)}.  We now have the exogenous variables for the selected unit.  This is called abduction and was computed based on actual measured data.

\subsection{Action: to intervene on counterfactual variables}
\label{action}
\begin{figure}[h!]
\centering
\begin{tikzpicture}
\Vertex[x=1,size = 1,label = $X$, fontscale = 2]{X}
\Vertex[size =1, x=5,label = $y$, fontscale =2.5]{Y}
\Vertex[size=1,x=3,y=2,fontscale = 2,label = $z$]{Z}
\Vertex[size=1,x=7,y=0,label = $U_y$,fontscale = 2]{Uy}
\Vertex[size=1,x=1,y=2,label = $U_z$,fontscale = 2]{Uz}
\Edge[Direct](Uy)(Y)
\Edge[Direct](Z)(X)
\Edge[Direct](Z)(Y)
\end{tikzpicture}
\caption{DAG showing common parent confounding but now with deleted edge between $U_z$ and \textit{z}, showing a counterfactual quantity independent of its' parents.}
\label{conf21}
\end{figure}
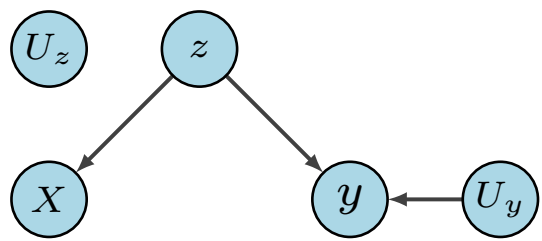

In the second step, called action, we input the new counterfactual features, for example,\textit{ Z=z}.  This means that \textit{Z} is no longer a function of \textit{Uz} and we have therefore deleted the arrow from \textit{Uz} to \textit{Z} and removed the relationship that \textit{Z} had with its parent, \textit{Uz}.  See Figure \ref{conf21}.  When we apply a counterfactual, we remove the relationship the feature has with its parents, regardless if they are endogenous or exogenous features. This is also called applying Pearl's do-operator to Z. This is a vital step in computing counterfactuals.  we now have a new modified (or surgically altered) model with which we compute counterfactual outcomes.

\subsection{Prediction}
\label{prediction}
In the final step, prediction, we use the new \textit{Z = z} feature value as well as the \textit{U} variables computed in step 1, and compute the new \textit{y = f(Z=z,Uy)}.  Detailed examples will be presented in the results of how to perform this using more complicated data.

\vspace{1cm}

\newpage
\section{Methodology}
\label{methods}
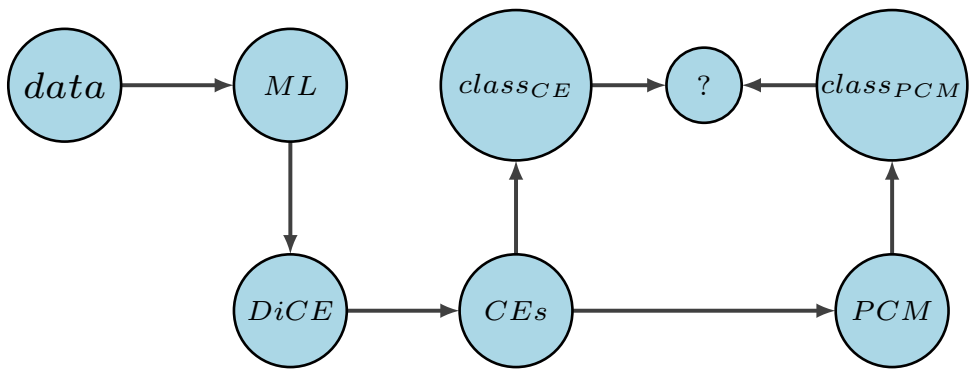
\begin{figure}[h!]
\centering
\begin{tikzpicture}
\Vertex[size=1.5,x=-4,y=3,label = $data$,fontscale = 2]{data}
\Vertex[size=1.5,x=-1,y=3,fontscale = 1.5,label = $ML$]{ML}
\Vertex[x=-1,label = $DiCE$, size=1.5,fontscale = 1.5]{DICE}
\Vertex[size =1.5, x=2,label = $CEs$, fontscale =1.5]{CE}
\Vertex[size=1.5,x=7,y=0,label = $PCM$,fontscale = 1.5]{PCM}
\Vertex[size=2,x=2,y=3,label = $class_{CE}$,fontscale = 1.5]{CLCE}
\Vertex[size=2,x=7,y=3,label = $class_{PCM}$,fontscale = 1.5]{CLPCM}
\Vertex[size=1,x=4.5,y=3,label = $?$,fontscale = 1.5]{Q}

\Edge[Direct](data)(ML)
\Edge[Direct](ML)(DICE)
\Edge[Direct](DICE)(CE)
\Edge[Direct](CE)(CLCE)
\Edge[Direct](CE)(PCM)
\Edge[Direct](PCM)(CLPCM)
\Edge[Direct](CLPCM)(Q)
\Edge[Direct](CLCE)(Q)
\end{tikzpicture}
\caption{Methodology flow generating DiCE CEs, feeding the CEs into PCM and comparing the output classes.}
\label{conf29}
\end{figure}

The CE method used in this study to generate CEs was DiCE ML \cite{Mothilal2020}. This was used because it is readily available as a Python package that is easy to implement.  The following presents the workflow for evaluating CEs using SCMs.  Also see Figure \ref{conf29}.

\begin{enumerate}
    \item Select a causal structure based on the three types discussed above and generate a dataset based on the SCM.
    \item Train a supervised learning model on the data.  This model is used in the DiCE CE generation process.
    \item Identify a case/unit in the dataset.
   
    \item Generate CEs using DiCE which utilizes the model trained in the earlier step. The CEs cause the output to switch to the opposite class.
     \item Now separately, carry out Abduction in Pearl's method from Section \ref{abduction}.  This is to compute all the exogenous features.
     \item Input DiCE CEs into the SCMs according to Pearl's Action method (Section \ref{action}).

     \item Estimate the output class using Pearl's Prediction method ((Section \ref{prediction}).

   
\end{enumerate}

\vspace{1cm}

\subsection{Experiments}

This section details the experiments based on the three causal structures above, but the datasets were made more complex with the aim of more closely modeling real life.  The dataset shown in Table \ref{dataset1} comprised seven features and was based very loosely on features describing a university student taking a course.  

\begin{table}[h!]
\centering
\caption{Description of features used in this study.}
\vspace{0.25cm}
\begin{tabular}{lll}
\hline

Feature & Description & Statistics \\
\hline
x1      & Grades      & $\mu$=50, $\sigma$=5       \\
x2      & Age         & $\mu$=20, $\sigma$=1       \\
x3      & Grades      & $\mu$=45, $\sigma$=6       \\
x4      & Gender      & \textit{p} = 0.6        \\
x5      & Bursary     &\textit{p} = 0.3        \\
x6      & Grades      & $\mu$=70, $\sigma$=5       \\
x7      & Grades      & $\mu$=50, $\sigma$=5     \\ \hline
\end{tabular}
\label{dataset1}
\end{table}

\subsection{Experiment 1: Chain causal structure}
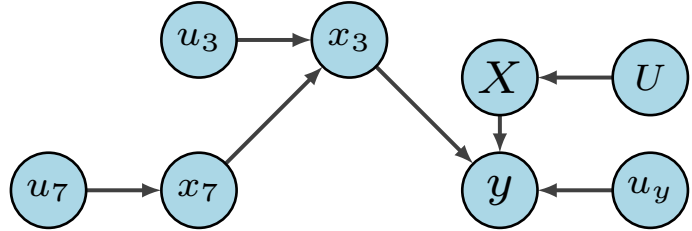
\begin{figure}[h!]
\centering
\begin{tikzpicture}
\Vertex[x=1,size = 1,y=2,label = $u_3$, fontscale = 2]{u3}
\Vertex[x=1,size = 1,label = $x_7$, fontscale = 2]{x7}
\Vertex[size =1, x=5,label = $y$, fontscale =2.5]{Y}
\Vertex[size =1, x=5,y=1.5,label = $X$, fontscale =2.5]{X}
\Vertex[size=1,x=3,y=2,fontscale = 2,label = $x_3$]{x3}
\Vertex[size=1,x=7,label = $u_y$,fontscale = 2]{Uy}
\Vertex[size=1,x=-1,label = $u_7$,fontscale = 2]{U7}
\Vertex[size=1,x=7,y = 1.5,label = $U$,fontscale = 2]{U}
\Edge[Direct](Uy)(Y)
\Edge[Direct](X)(Y)
\Edge[Direct](U)(X)
\Edge[Direct](u3)(x3)
\Edge[Direct](U7)(x7)
\Edge[Direct](x7)(x3)
\Edge[Direct](x3)(Y)
\end{tikzpicture}
\caption{Causal structure for experiment 1 that is based on a chain (mediator) causal structure.}
\label{exp100}
\end{figure}

The DAG and SCMs for experiment 1 are based on the chain causal structure (Section \ref{causalstructurechain}) and is shown in Figure \ref{exp100} and Equations \ref{exp100a} to \ref{exp100b}.  Note that in the SCM, y is not a function of $x_7$ and would not make use of it in predictions; however an ML model will include $x_7$ in the training and prediction.   In the DAG, for brevity and space-saving, note that X refers to all the features not represented by $x_7$ and $x_3$, namely $x_1$, $x_2$, $x_4$, $x_5$, $x_6$.  They are features feeding directly into y and are considered parents of y. Also, \textit{U} refers to all the exogenous variables feeding into \textit{X}, namely $U_1$, $U_2$, $U_4$, $U_5$, $U_6$.  In the SCMs, the $\beta$ coefficients from 0 to 8 were arbitrarily selected as follows: 0.4, 0.6, 0.4, 0.6, 0.7, 0.4, 0.4, 0.3, 0.7.

\vspace{2cm}

\begin{align}  
&x_1 = u_1 \label{exp100a}\\
&x_2 = u_2\\
&x_3= \beta_1 x_7 + u_3  \\
&x_4 = u_4\\
&x_5 = u_5\\
&x_6 = u_6\\
&x_7 = u_7\\
&y = \beta_0 + \beta_1 x_1 + \beta_2 x_2 + \beta_3 x_3 + \beta_4 x_4 + \beta_5 x_5 + \beta_6 x_6 + U_y  \label{exp100b}
\end{align}

\subsubsection{Experiment 2:  Fork (common cause) causal structure}
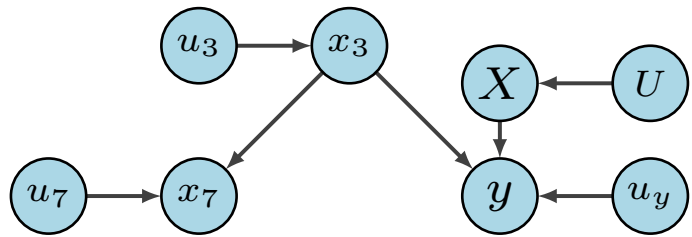
\begin{figure}[h!]
\centering
\begin{tikzpicture}
\Vertex[x=1,size = 1,y=2,label = $u_3$, fontscale = 2]{u3}
\Vertex[x=1,size = 1,label = $x_7$, fontscale = 2]{x7}
\Vertex[size =1, x=5,label = $y$, fontscale =2.5]{Y}
\Vertex[size =1, x=5,y=1.5,label = $X$, fontscale =2.5]{X}
\Vertex[size=1,x=3,y=2,fontscale = 2,label = $x_3$]{x3}
\Vertex[size=1,x=7,label = $u_y$,fontscale = 2]{Uy}
\Vertex[size=1,x=-1,label = $u_7$,fontscale = 2]{U7}
\Vertex[size=1,x=7,y = 1.5,label = $U$,fontscale = 2]{U}
\Edge[Direct](Uy)(Y)
\Edge[Direct](X)(Y)
\Edge[Direct](U)(X)
\Edge[Direct](u3)(x3)
\Edge[Direct](U7)(x7)
\Edge[Direct](x3)(x7)
\Edge[Direct](x3)(Y)
\end{tikzpicture}
\caption{Causal structure for experiment 2 that is based on a fork (common cause/parent).}
\label{exp1000}
\end{figure}

The second experiment was based on the fork structure from Section \ref{common parent- section}.  The DAG and SCMs are presented in Figure \ref{exp1000} and Equations \ref{exp200a} and \ref{exp200b}.  Whereas the causal path in Experiment 1 flowed from $x_7$ to $x_3$ to y,  in Experiment 2, $x_3$ is the cause of both $x_7$ and y.  Everything else is the same as Experiment 1.

\vspace{2cm}

\begin{align}  
&x_1 = u_1 \label{exp200a}\\
&x_2 = u_2\\
&x_3 = u_3\\
&x_4 = u_4\\
&x_5 = u_5\\
&x_6 = u_6\\
&x_7= \beta_8 x_3 + u_7 \\ 
&y = \beta_0 + \beta_1 x_1 + \beta_2 x_2 + \beta_3 x_3 + \beta4 x_4 + \beta5 x_5 + \beta_6 x_6 + u_y  \label{exp200b}
\end{align}

\subsection{Experiment 3:  Collider causal structure}

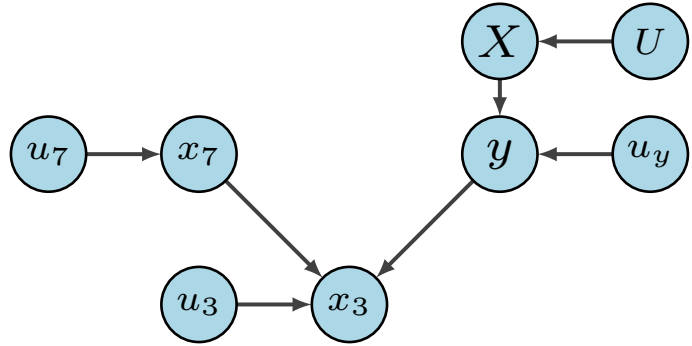
\begin{figure}[h!]
\centering
\begin{tikzpicture}
\Vertex[x=1,size = 1,y=-2,label = $u_3$, fontscale = 2]{u3}
\Vertex[x=1,size = 1,label = $x_7$, fontscale = 2]{x7}
\Vertex[size =1, x=5,label = $y$, fontscale =2.5]{Y}
\Vertex[size =1, x=5,y=1.5,label = $X$, fontscale =2.5]{X}
\Vertex[size=1,x=3,y=-2,fontscale = 2,label = $x_3$]{x3}
\Vertex[size=1,x=7,label = $u_y$,fontscale = 2]{Uy}
\Vertex[size=1,x=-1,label = $u_7$,fontscale = 2]{U7}
\Vertex[size=1,x=7,y = 1.5,label = $U$,fontscale = 2]{U}
\Edge[Direct](Uy)(Y)
\Edge[Direct](X)(Y)
\Edge[Direct](U)(X)
\Edge[Direct](u3)(x3)
\Edge[Direct](U7)(x7)
\Edge[Direct](x7)(x3)
\Edge[Direct](Y)(x3)
\end{tikzpicture}
\caption{Causal structure for experiment 3 that is based a collider causal structure.}
\label{exp3000}
\end{figure}

The DAG and SCMs for experiment 3 based on the collider causal structure shown in Figure \ref{exp3000} and Equations \ref{451} to \ref{452}.  Note again that there is no causal path directly between $x_7$ and y and $x_3$ is not a parent of y.  Therefore the SCM does not consider $x_7$ nor $x_3$ as a parent of y and is not included in computing y (see Equation \ref{452}). However, in an ML model, $x_7$ and $x_3$ would most likely be included.  This again highlights the distinction between ML modelling and SCM modelling.

\vspace{1cm}

\begin{align}  
&x_1 = u_1  \label{451}\\
&x_2 = u_2\\
&x_3= \beta_1 x_7 + u_3 \\
&x_4 = u_4\\
&x_5 = u_5\\
&x_6 = u_6\\
&x_7 = u_7\\
&y = \beta_0 + \beta_1 x_1 + \beta_2 x_2 + \beta_4 x_4 + \beta_5 x_5 + \beta_6 x_6 + u_y  \label{452}
\end{align}

\subsection{Limitations and Assumptions}
In all the CE computations and evaluations in this study, I make the assumption that all features are actionable and all counterfactual values are plausible.  The aim of this study was not to study actionability and plausibility but rather causality. In real life, however, it is highly unlikely that all features are actionable and all counterfactuals are plausible.  This was assumed to simplify the study.

\section{Results}
\label{results}

This section presents two main results.  The first is to show the PCM \textbf{method} using CEs from DiCE (Section \ref{results:PCM}).  The second is to show the \textbf{results} of this method on 30 examples from the three causal structures (Section \ref{results:causalstruc}).


\subsection{Pearl's Counterfactual Method (PCM)}
\label{results:PCM}
\subsubsection{Experiment 1: Chain causal structure}
\label{results:exp1}
\begin{figure}[h!]
    \centering
    \includegraphics[width = 130mm]{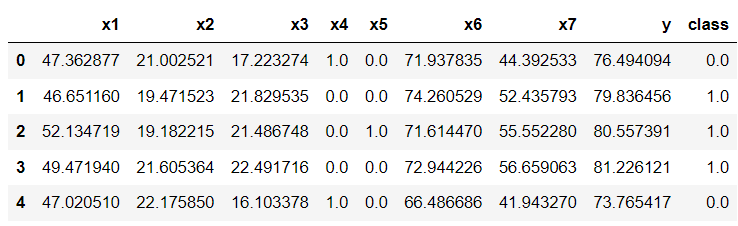}
    \caption{First five rows of the data generated in step 1.}
    \label{fig:mediatorDataset}
\end{figure}

Following the steps outlined under Section \ref{methods}, we begin by:

\paragraph{Step 1:}
generating a dataset by sampling each feature in Table \ref{dataset1} and generating the data according to the SCMs of Equations \ref{exp100a} to \ref{exp100b}.  A sample of the data generated is shown in the first five rows in Figure \ref{fig:mediatorDataset}. Note that y is a numeric value and class is a binary 0 or 1.  DiCE only operates on changing class values so I converted the output y to a class by making any value greater or equal to the mean value, a one, and anything below the mean value, a zero.  This was the same in all subsequent results.

\paragraph{Step 2:}
Next, train a logistic regression model on the data which is used in the DiCE operations.  Logistic regression is used because the output is a class.  Any supervised learning model that performs classification can be used here.

\begin{figure}[h!]
    \centering
    \includegraphics[width = 130mm]{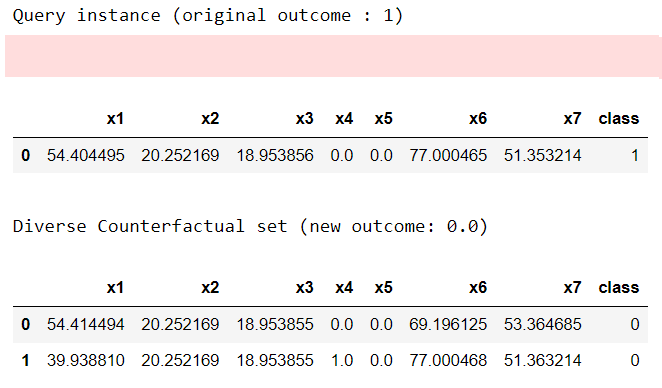}
    \caption{Random unit selected for experiment 1, shown at the top. In this DiCE computation, two DiCE counterfactual explanations were generated, shown by the two rows at the bottom.}
    \label{fig:mediator}
\end{figure}

\paragraph{Step 3:}
Select any unit in the dataset that you are interested in, on which to compute counterfactuals.  This unit is presented in Figure \ref{fig:mediator}, together with two CEs computed by DiCE.  Notice the original class was 1 and the counterfactuals switched the class to 0.

\newpage
\paragraph{Steps 4 and 5:}
Next we carry out Abduction, which is to compute the exogenous variables by using the actual feature values for this unit shown in Figure \ref{fig:mediator}. Using the values from the top row in Figure \ref{fig:mediator}, I reproduce Equations \ref{451} to \ref{452} but now with values inserted to compute the \textit{U} values which are shown in Table \ref{tab:exp1exog}.

\begin{align}  
&54.5 = \boldsymbol{u_1}  \label{551}\\
&20.2 = \boldsymbol{u_2}\\
&18.9= 0.6\cdot53.3 +\boldsymbol{u_3} \\
&0.0~~ = \boldsymbol{u_4}\\
&0.0~~ = \boldsymbol{u_5}\\
&69.2 = \boldsymbol{u_6}\\
&53.3 =\boldsymbol{ u_7}\\
&81.7 = 0.4 + 0.6 \cdot 54.5 + 0.4\cdot20.2 + 0.6\cdot18.9 + 0.7\cdot0 + 0.4\cdot0 + 0.4\cdot77.0 + \boldsymbol{u_y} \label{552}
\end{align}

\begin{table}[h!]
    \centering
    \caption{Exogenous variables for experiment 1.}
    \begin{tabular}{cccccccc} \hline
        $u_1$ & $u_2$ & $u_3$ & $u_4$ & $u_5$ & $u_6$ & $u_7$ & $u_y$ \\ \hline
         54.4&20.2& -1.58 &0&0&77.0&51.4&-1.58\\
         \hline
    \end{tabular}
    
    \label{tab:exp1exog}
\end{table}

\paragraph{Step 6:}
 In this example I select counterfactual 0: i.e. the first row of counterfactuals in Figure \ref{fig:mediator}.  Here we can see that two features, $x_6= 69.2$ and $x_7 = 53.4$ have been changed by DiCE in order to obtain a different output class of 0.  The rest remain unchanged. Therefore when applying Pearl's method, Action, if the feature is different, we delete its' parents and input the new counterfactual value. This is because when we intervene on a feature, we now introduce a new mechanism that determines the state of those features, and it no longer ``listens" to it's parents \cite{Pearl2017}.

We now apply the above (i.e the exogenous variables and the CEs) to the SCMs for this experiment shown below.  Ultimately, we are trying to compute the output y from the SCM and compare it with the output y from the DiCE CEs.

\begin{align}  
&x_1 = u_1 = 54.4 \label{750}\\
&x_2 = u_2 = 20.2\\
&x_3= 0.6\boldsymbol{\cdot} 53.3 + (-1.58) \\
&x_4 = u_4 = 0.0\\
&x_5 = u_5 = 0.0\\
&x_6 = CE_6 = 69.2\\
&x_7 = CE_7 = 53.3\\
&y = \beta_0 + \beta_1 u_1 + \beta_2 u_2 + \beta_3 x_3 + \beta_4 u_4 + \beta_5 u_5 + \beta_6 CE_6 + u_y 
   =  79.08
\label{751}
\end{align}

The output y = 79.08 is greater than the mean of 79.07. Therefore when applying the CEs computed by DiCE to PCM, we obtain a class of 1, whereas DiCE for this unit, predicted a class of 0.   

\subsubsection{Experiment 2 -  Fork causal structure}
\label{results:exp2}
\begin{figure}[h!]
    \centering
    \includegraphics[width = 130mm]{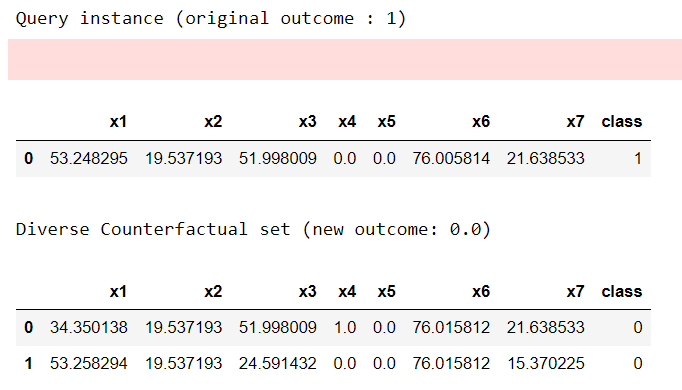}
    \caption{Selected unit and two DiCE CEs generated for the fork causal structure.}
    \label{fig:ForkDiCE}
\end{figure}

Following the same steps as above, for the fork causal structure, we generate the dataset (Step 1), train a classification model (Step 2), select a unit/case and generate DiCE CEs (Steps 3 and 4, see Figure \ref{fig:ForkDiCE}). compute exogeneous variables (Steps 4 and 5), identify counterfactual values and apply Action (Step 6), and finally compute the SCM output and compare with DiCE output (Step 7).

Figure \ref{fig:ForkDiCE} shows the unit that was selected as well as the two DiCE CEs.

Table \ref{tab:exp2exog} shows the calculated exogenous variables.

\begin{table}[h!]
    \centering
    \caption{Exogenous variables for experiment 2.}
    \begin{tabular}{cccccccc} \hline
        $u_1$ & $u_2$ & $u_3$ & $u_4$ & $u_5$ & $u_6$ & $u_7$ & $u_y$ \\ \hline
         53.2&19.5& 51.9 &0&0&76.0&0.84&0.84\\
         \hline
    \end{tabular}
    
    \label{tab:exp2exog}
\end{table}

If we again use $CE_0$ in our example, we can see from Figure \ref{fig:ForkDiCE}, that features $x_1 = 34.4$ and $x_4 = 1.0$ are different from the original: they are counterfactuals.  We therefore pay special attention to them and delete their original causal mechanism and replace it with the new values; the rest of the features remain unchanged in the SCMs.

\begin{align}  
&y = \beta_0 + \beta_1\cdot CE_1 + \beta_2 u_2 + \beta_3 u_3 + \beta4 \cdot CE_4 + \beta_5 u_5 + \beta_6 u_6 + u_y  = 91.9 \label{exp210}
\end{align}

The output y = 91.6 is smaller than the mean of 93.6. Therefore when applying the CEs computed by DiCE to PCM, we obtain a class of 0 which is the same class predicted by DiCE.

\subsubsection{Experiment 3 - Collider causal structure}
\label{results:exp3}
\begin{figure}[h!]
    \centering
    \includegraphics[width = 130mm]{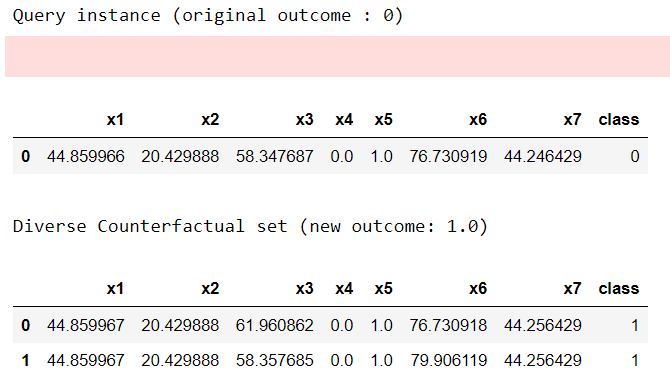}
    \caption{Selected unit for Experiment 3 with two DiCE CEs generated for the collider causal structure.}
    \label{fig:CollDiCE}
\end{figure}

The final example is for the collider causal structure and we again follow the same six steps as before. We will not repeat the details here, only the results.  Figure \ref{fig:CollDiCE} shows the unit selected and the DiCE counterfactuals and Table \ref{tab:exp3exog} shows the computed \textit{U} variables.

\begin{table}[h!]
    \centering
    \caption{Exogenous variables for experiment 3.}
    \begin{tabular}{cccccccc} \hline
        $u_1$ & $u_2$ & $u_3$ & $u_4$ & $u_5$ & $u_6$ & $u_7$ & $u_y$ \\ \hline
         44.8&20.4& 0.43 &0&1&76.7&44.3&0.43\\
         \hline
    \end{tabular}
    
    \label{tab:exp3exog}
\end{table}

If we again use $CE_0$ in this example from Figure \ref{fig:CollDiCE}, we can see that only features $x_3 = 61.9$ is different from the original.  However, due to the causal structure of the collider, $x_3$ is not a parent of y and is therefore ignored.  The output y is therefore calculated as follows:

\begin{align}  
&y = \beta_0 + \beta_1 u_1 + \beta_2 u_2 + \beta_4 u_4 + \beta_5 u_5 + \beta_6 u_6 + u_y  = 67.0 \label{exp211}
\end{align}

The output y = 67.0 is greater than the mean of 66.84. Therefore when applying the CEs computed by DiCE to PCM, we obtain a class of 1 which is the same class predicted by DiCE.

\subsection{Effect of Causal Structures}
\label{results:causalstruc}
Section \ref{results:PCM} presented the step by step method for computing CEs via DiCE and then using these values in PCM to compute counterfactual outputs.  The aim was to present the method of using PCM.  This section, still using the same method, aims to obtain an idea of whether different causal structures produce different counterfactual outcomes using PCM.  We do this by performing this method on thirty DiCE CEs; ten for each causal structure.  

\subsubsection{Chain (mediator)}
Table \ref{results:tab1} shows the results of selecting 5 units, with DiCE generating two counterfactual explanations for each unit, totalling ten CEs.  Each CE was input into the chain SCMs and the 6 steps were followed as above.  The column ``class" refers to the class of the original unit and the CE classes.  The column ``class (PCM)" is the class predicted using Pearl's counterfactual method.  We can see that three out of the ten times, PCM predicted a different class than the DiCE method.  This is highlighted in gray in the table.

\begin{table}[h!]
    \centering
    \caption{Results for chain causal structure}
    \begin{tabular}{l|ccccccccc|cc} \hline 
   \textbf{Item}& $x_1$ & $x_2$ & $x_3$ & $x_4$ & $x_5$ & $x_6$ & $x_7$ & y & class & class (PCM) & y (PCM)\\\hline \hline
    \textbf{\textit{Unit 1}} & 54.4& 20.2&18.9 &0 &0 &77.0 &51.3 & 81.7 & 1 &-\\ \hline
      \textbf{\textit{U}}& 54.4&20.2& -1.58 &0&0&77.0&51.4&-1.58&-&-\\
    \rowcolor{Gray} $CE_0$& 54.4& 20.2&18.9 &0 &0 &\textbf{69.2} &\textbf{53.4} & - & 0 & 1 & 79.1\\
      $CE_1$& \textbf{39.9}& 20.2&18.9 &\textbf{1} &0 &77.0 &51.3 & - & 0 & 0 & 77.8\\
         \hline\hline

         \textbf{\textit{Unit 2}} & 52.5& 18.7&15.9 &0 &0 &80.5 &40.1 & 80.9 & 1 &-\\ \hline
      \textbf{\textit{U}}& 52.5& 18.7&-0.16 &0 &0 &80.5 &40.1 & -0.16 & - &-\\
      $CE_0$& 52.5& 18.7&15.9 &0 &0 &\textbf{73.1} &40.1 & - & 0 &0 & 77.5\\
      $CE_1$& \textbf{41.4}& 18.7&15.9 &0 &0 &\textbf{70.9} &40.1 & - & 0 & 0 & 70.4\\
         \hline\hline

         \textbf{\textit{Unit 3}} & 57.5& 21.3&23.2 &0 &1 &61.9 &58.6 & 82.3 & 1 &-\\ \hline
      \textbf{\textit{U}}& 57.5& 21.3&-0.2 &0 &1 &61.9 &58.6 & -0.2 & - &-\\
      \rowcolor{Gray}  $CE_0$& 57.5& 21.3&\textbf{19.0} &0 &1 &61.9 &58.6 & - & 0 & 1&79.8\\
      $CE_1$& 57.5& 21.3&\textbf{15.1} &0 &\textbf{0} &61.9 &58.6 & - & 0 & 0 &77.1\\
         \hline\hline

         \textbf{\textit{Unit 4}} & 54.6& 20.2&19.9 &0 &0 &66.2 &54.1 & 77.9& 0 &-\\ \hline
      \textbf{\textit{U}}& 54.6& 20.2&-1.7 &0 &0 &66.2 &54.1 & -1.7 & - &-\\
      $CE_0$& \textbf{62.7}& 20.2&19.9 &0 &0 &66.2 &54.1  & - & 1 & 1&82.8\\
      $CE_1$& \textbf{62.7}& \textbf{18.5}&19.9 &0 &0 &66.2 &54.1  & - & 1 & 1 &82.1\\
         \hline\hline

         \textbf{\textit{Unit 5}} & 50.4& 20.4&25.6 &0 &1 &67.5 &58.8 & 83.7& 1 &-\\ \hline
      \textbf{\textit{U}}&50.4& 20.4&2.1 &0 &1 &67.5 &58.8  & 2.1 & - &-\\
       \rowcolor{Gray} $CE_0$& 50.4& 20.4&\textbf{21.7} &0 &1 &67.5 &58.8 & - & 0 & 1&80.9\\
      $CE_1$& \textbf{35.3}& 20.4&25.6 &0 &1 &67.5 &58.8  & - & 0 & 0 &74.2\\
         \hline\hline
    \end{tabular}
    
    \label{results:tab1}
\end{table}

\subsubsection{Fork causal structure}

Table \ref{results:tab2} presents the results for ten CEs using the fork causal structure. The gray rows show where the PCM method generates the opposite class to DiCE.  This occurred two out of the ten times.
\begin{table}[h!]
    \centering
    \caption{Results for fork causal structure. }
    \begin{tabular}{l|ccccccccc|cc} \hline 
   \textbf{Item}& $x_1$ & $x_2$ & $x_3$ & $x_4$ & $x_5$ & $x_6$ & $x_7$ & y & class & class (PCM) & y (PCM)\\\hline \hline
    \textbf{\textit{Unit 1}} & 47.5& 19.53&50.6 &0 &0 &68.5 &19.9 & 94.1 & 0 &-\\ \hline
      \textbf{\textit{U}}& 47.5& 19.53&50.6 &0 &0 &68.5 &-0.3&-0.3&-&-\\
   $CE_0$& 47.5& 19.53&\textbf{51.7} &\textbf{1 }&0 &68.5 &19.9  & - & 1 & 1 & 95.5\\
      $CE_1$& 47.5& 19.53&\textbf{55.8 }&0 &0 &68.5 &19.9  & - & 1 & 1 & 97.0\\
         \hline\hline

         \textbf{\textit{Unit 2}} &56.7& 18.9&42.4 &1 &0 &63.9 &15.9 & 92.6 & 1 &-\\ \hline
      \textbf{\textit{U}}&56.7& 18.9&42.4 &1 &0 &63.9 &-1.11 & -1.11 & - &-\\
      $CE_0$& \textbf{51.2}& \textbf{20.5}&42.4 &1 &0 &63.9 &15.9  & - & 0 &0 & 89.9\\
      $CE_1$& \textbf{62.4}& 18.9&\textbf{28.2} &1 &0 &63.9 &15.9  & - & 0 & 0 & 87.5\\
         \hline\hline

         \textbf{\textit{Unit 3}} & 52.3& 18.7&43.1 &1 &0 &64.4 &18.6 & 92.9 & 1&-\\ \hline
      \textbf{\textit{U}}& 52.3& 18.7&43.1 &1 &0 &64.4 &1.36 & 1.36 & - &-\\
      \rowcolor{Gray} $CE_0$& \textbf{53.5}& \textbf{22.1}&43.1 &1 &0 &64.4 &18.6 & - & 0 & 1&95.0\\
       $CE_1$& 52.3& \textbf{20.3}&43.1 &1 &0 &64.4 &18.6 & - & 0 & 0 &93.6\\
         \hline\hline

         \textbf{\textit{Unit 4}} & 46.6& 20.2&52.9 &1 &0 &72.2 &20.4 & 96.6& 1 &-\\ \hline
      \textbf{\textit{U}}& 46.6& 20.2&52.9 &1 &0 &72.2 &-0.8& -0.8& - &-\\
      $CE_0$&\textbf{36.4}& \textbf{22.3}&52.9 &1 &0 &72.2 &20.4 & - & 0 & 0&91.7\\
      $CE_1$& \textbf{38.8}& 20.2&52.9 &1 &0 &72.2 &20.4  & - & 0 & 0 &92.3\\
         \hline\hline

         \textbf{\textit{Unit 5}} & 54.4& 21.1&46.7&0 &0 &74.9 &19.9 & 100.7& 1 &-\\ \hline
      \textbf{\textit{U}}& 54.4& 21.1&46.7&0 &0 &74.9 &19.9 & 2.1 & - &-\\
       $CE_0$&  \textbf{36.4}& 21.1&46.7&0 &0 &\textbf{70.1} &19.9& - & 0 & 0&87.9\\
      \rowcolor{Gray} $CE_1$&  \textbf{46.1}& 21.1&46.7&0 &0 &74.9 &19.9 & - & 0 & 1 &95.7\\
         \hline\hline
    \end{tabular}
    
    \label{results:tab2}
\end{table}

\subsubsection{Collider causal structure}
Table \ref{results:tab3} presents the PCM results for ten CEs generated from the collider causal structure.  We can see here that five out of ten (50\%) of the DiCE CEs produced outputs that conflicted with the PCM output classes.   These are shown in gray.

\begin{table}[th]
    \centering
    \caption{Results for collider causal structure}
    \begin{tabular}{l|ccccccccc|cc} \hline 
   \textbf{Item}& $x_1$ & $x_2$ & $x_3$ & $x_4$ & $x_5$ & $x_6$ & $x_7$ & y & class & class (PCM) & y (PCM)\\\hline \hline
    \textbf{\textit{Unit 1}} & 47.3& 21.1&57.4 &1 &0 &59.7 &48.9 &62.3 & 0 &-\\ \hline
      \textbf{\textit{U}}&  47.3& 21.1&0.48 &1 &0 &59.7 &48.9 &0.48&-&-\\
    $CE_0$&  47.3& 21.1&57.4 &1 &0 &\textbf{85.2} &48.9  & - &1& 1 & 72.5\\
      \rowcolor{Gray} $CE_1$&  47.3& \textbf{20.7}&\textbf{69.5} &1 &0 &59.7 &48.9  & - & 1 & 0 & 62.1\\
         \hline\hline

         \textbf{\textit{Unit 2}} & 46.8& 19.7&58.9 &0 &1 &68.9 &51.9 & 64.0 & 0 &-\\ \hline
      \textbf{\textit{U}}& 46.8& 19.7&-0.3&0 &1 &68.9 &51.9& -0.3 & - &-\\
    \rowcolor{Gray}  $CE_0$& 46.8& 19.7&\textbf{69.7} &0 &1 &68.9 &51.9 & - & 1 &0 & 64.05\\
       $CE_1$&\textbf{61.4}& 19.7&58.9 &0 &1 &68.9 &51.9 & - & 1 & 1 & 72.7\\
         \hline\hline

         \textbf{\textit{Unit 3}} & 43.3& 19.4&56.7 &0 &0 &75.3 &51.4 & 62.7 & 0 &-\\ \hline
      \textbf{\textit{U}}& 43.3& 19.4&-1.5 &0 &0 &75.3 &51.4 & -1.5 & - &-\\
      \rowcolor{Gray}  $CE_0$& \textbf{46.9}& 19.4&\textbf{63.1} &0 &0 &75.3 &51.4  & - & 1 & 0&64.8\\
      $CE_1$& \textbf{60.5}&\textbf{ 16.8}&56.7 &0 &0 &75.3 &51.4  & - & 1 & 1&72.0\\
         \hline\hline

         \textbf{\textit{Unit 4}} & 45.5&19.9&63.8 &0 &1 &66.3 &59.7 & 64.1& 0 &-\\ \hline
      \textbf{\textit{U}}& 45.5&19.9&1.5&0 &1 &66.3 &59.7  & 1.5 & - &-\\
      $CE_0$& 45.5&19.9&63.8 &0 &1 &66.3 &47.17  & - & 1 & 1&73.1\\
      $CE_1$& 45.5&19.9&63.8 &0 &1 &66.3 &48.6   & - & 1 & 1 &73.1\\
         \hline\hline

         \textbf{\textit{Unit 5}} & 43.6& 20.4&63.9 &0 &1 &77.3 &55.3 & 67.4& 1 &-\\ \hline
      \textbf{\textit{U}}&43.6& 20.4&1.3 &0 &1 &77.3 &55.3& 1.3 & - &-\\
       \rowcolor{Gray} $CE_0$& 43.6& 20.4&\textbf{54.8} &0 &1 &77.3 &55.3 & - & 0 & 1&67.4\\
     \rowcolor{Gray} $CE_1$& 43.6& 20.4&\textbf{61.5} &0 &\textbf{0} &77.3 &55.3  & - & 0 & 1 &67.4\\
         \hline\hline
    \end{tabular}
    
    \label{results:tab3}
\end{table}

\newpage
\section{Discussion}
\label{discussion}

This section discusses the results and specifically how this method compares and fits in with current literature.

\subsection{Pearl's Counterfactual Method}
Verma et al. \cite{Verma2021} identified a current challenge in the CE literature being the lack of using causal models (SCMs) to guide the counterfactual explanations. Mahajan et al. \cite{Mahajan2020} stated that feasibility (i.e. coming from a possible world) in CEs is fundamentally a causal concept and that it is important to preserve causal relationships.  That is, if we vary one feature, we have to consider how other features are causally related to that feature.  They give an example where a CE might recommend increasing education level to masters to obtain a different desired outcome.  According to them, if we change education level, then the age of the person also needs to change due to being causally linked with education level.  To preserve causal constraints, they proposed to constrain the distance between features based on an SCM.  Riccardo Guidotti \cite{guidotti22} also mentions causality as a requirement for CEs: for a counterfactual explanation to be plausible and actionable, it should maintain causal relationships between features.  Both of these papers make sense.  We want to use the true causal structure in the data.  This is even a main contribution of this paper. However, when dealing with counterfactuals, there is some nuance.  According to Pearl, when we intervene on a feature and perform do-calculus, we manipulate and change the original causal structure in the data by \textbf{deleting the relationship} between that feature and its' parents.  This is slightly different to what Mahajan et al. \cite{Mahajan2020} and Guidotti \cite{guidotti22} are requiring. Indeed, we want to make use of the true causal structure in the data, however counterfactual features are no longer constrained by their parents as described above in Pearl's Action method (see Section \ref{action}).  Therefore if we apply Pearl's method to Mahajan et al's example, if age is a parent of education level in the original causal structure, and the CE has recommended increasing education level, then because we are now intervening on education, we free it from the influence of its surrounding and it becomes only a function of our intervention. Age is no longer a cause of education and these two features are no longer causally constrained.  However, this differs slightly from Mahajan et al. who aim to preserve the causal relationship.
 
 Furthermore, by preserving the causal structure of the counterfactual variable, are we not simply maintaining the same distribution of data?  Recall that counterfactuals are by nature not in the distribution.  Of course, our features must still be feasible; i.e. they must come from the real world.  However, it seems that maintaining the causal structure between a counterfactual and its parents keeps the problem on Rung 1 of the Pearl causal hierarchy: association \cite{Pearl2017}.  A counterfactual by definition is to keep everything else the same, and only change \textit{that} feature.  If other things are changing with it according to the original data generation process, then it seems that it is not a counterfactual. 
 I acknowledge however, that if I have misunderstood the above mentioned literature, then I am open for correction.

The reader may then ask why am I making such a fuss over using SCMs if it seems that I don't care for the constraints in the causal structure.  Recall that it is \textbf{only the counterfactual feature that is deleted from its' parents}; the other input features maintain their causal structure.  Furthermore, the causal relationship between the counterfactual feature and the output remains untouched.   
Indeed, we must preserve feasibility by keeping the data in the real world, but we cannot do this by preserving a counterfactual feature’s causal relationship with its’ parents.


\subsection{Effect of Causal Structures}
The second aim was to see, using PCM, how different causal structures affected the CEs generated using DiCE.  We see that first of all, out of 30 different results, 10 (33\%) showed conflict between the CE output and the PCM output.  Although definitely not conclusive, the results suggest that DiCE CEs produce incorrect estimates (assuming PCM is correct). For more conclusive results, I suggest performing a simulation study of 500 to 1000 estimates.  The challenge here was that, at least for now, each PCM calculation was carried out by hand and therefore very challenging to perform hundreds of times.   Future work would develop code to simulate this type of study.

Another important result was that the collider causal structure counterfactuals showed 50\% conflict with the DiCE CEs.  Chain had 3/10 conflicts, fork had 2/10 and collider, 5/10 conflicts.  Again, these are not conclusive results, but they suggest that the type of causal structure may have a significant effect on the true counterfactual explanations generated.  Whereas ML models simply include all features in the model, the true SCM would not include all the features.  Actually, to include a collider in a statistical predictive model produces unwanted bias in the causal estimates \cite{Luque-Fernandes2018} that can ``sabotage a causal analysis"\cite{Morgan2007}.  These results suggest that simply training a model on the entire dataset may introduce unwanted error into our counterfactual estimates.  \textit{It is vital that we understand the underlying causal structure before blindly generating CEs.}
\subsection{Causal Discovery}
This study used simulated data with known ground truth causal structures.   The reader may argue that this method only works with simulated data and how can we validate CEs using PCM if we have real life data and don't know the causal structure?  This is where causal discovery comes in which refers to discovering the true causal structure in real life data \cite{Glymour2019}.  More explanation of this method is outside the scope of this study.  However, I believe that for us to have significantly more trust in CEs from DiCE and others, we must first perform causal discovery on our data in order to determine as best we can, the true causal structure.  We then perform CE validation by feeding the CEs into the PCM method on the true structure.  Real life datasets may be substantially more complicated that the ones I created in this study.  This is left for future work.

\subsection{Concluding Remarks and Future Work}
In the beginning of the article I argued that it is vital to evaluate (validate) counterfactual explanations (CEs).  A main reason is that CEs are generated from machine learning models that do not necessarily take into account the causal structure in the data.  This could lead to errors in the predictions and estimations of counterfactuals.  CEs are gaining much interest in the data community and if we don't deeply understand how we are generating CEs, then at best they will add no value.   

The results of this study showed that there was some conflict between the ML method for generating CEs, and Pearl's method.  This casts some doubt on how much we can trust CEs that are blindly generated by machine learning models.  Different causal structures showed different levels of conflict.  

Where do we go from here?  It is vital that before we generate CEs, we do our best to know the true causal structure and modify our models accordingly.  Causal discovery is essential for the future of CEs.  Future work would include carrying out PCM using CEs on real life data with a known causal structure.

I also believe that the ultimate way of validating CEs is to implement them on real data and measure the true output after a period of time.  This is of course, highly challenging. 

For more confidence in these results, it is best to simulate on hundreds of CEs and also apply it to real life data.

\bibliographystyle{unsrt} 
\bibliography{ReferencesJan23.bib}

\end{document}